\documentclass[runningheads]{llncs}

\usepackage[T1]{fontenc}
\usepackage{newtxtext}
\usepackage[varvw]{newtxmath}

\usepackage{graphicx}
\usepackage{booktabs}
\usepackage{amsmath}
\usepackage{url}
\usepackage{microtype}
\usepackage{algorithm}
\usepackage{algpseudocode}

\usepackage{pgfplots}
\usepgfplotslibrary{groupplots}
\pgfplotsset{compat=1.18}

\begin{document}

\title{Multi-Objective Evolutionary Semi-Supervised Learning for Resource-Aware Smart City Classification}
\titlerunning{MOEA-SSL for Resource-Aware Smart City Classification}

% TODO: Add/remove co-authors before submission.
\author{Francisco José Sedeño\inst{1}\orcidID{0009-0000-7797-3562} \and \\Jamal Toutouh\inst{1}\orcidID{0000-0003-1152-0346}}
\authorrunning{Sedeño and Toutouh}
\institute{ITIS Software, University of Malaga, Malaga, Spain\\
\email{\{sedeno,jamal\}@uma.es}}

\maketitle

\begin{abstract}
%Smart-city applications often combine scarce labeled data with abundant unlabeled observations and redundant measurements. This article presents MOEA-SSL, a multi-objective evolutionary approach for semi-supervised tabular classification that simultaneously optimizes predictive quality (measured as robust MacroF1) and the number of selected input features. A multi-objective evolutionary algorithm is applied to obtain different compromise solutions. Experimental evaluation on three Smart City classification datasets, considering 1\%, 5\%, and 10\% labeled data, shows that MOEA-SSL achieves competitive predictive performance compared with state-of-the-art methods while significantly reducing required input features in all evaluated conditions. The computed Pareto fronts also provide additional resource-efficient alternatives with limited degradation in classification performance.

Smart-city applications often combine scarce labeled data with abundant unlabeled observations. Semi-supervised learning (SSL) exploits both labeled and unlabeled data to improve classification under limited supervision. This article presents MOEA-SSL, a multi-objective evolutionary approach for lightweight semi-supervised tabular classification that jointly optimizes predictive quality and feature usage. Experimental evaluation on three Smart City classification datasets, considering 1\%, 5\%, and 10\% labeled data, shows that MOEA-SSL achieves competitive predictive performance compared with established lightweight SSL methods while significantly reducing feature usage relative to EA-SSL in all evaluated conditions. The computed Pareto fronts also provide lower-feature-use alternatives with limited degradation in classification performance.

%Smart-city systems often combine abundant unlabeled observations with scarce labels and redundant measurements. This article presents MOEA-SSL, a multi-objective evolutionary approach for semi-supervised tabular classification that extends the mono-objective EA-SSL formulation with NSGA-II. The method jointly minimizes negative robust MacroF1 and the fraction of distinct input features used by two evolved SSL views, producing Pareto alternatives between predictive quality and measurement complexity. Experiments on three Smart City binary classification datasets consider labeled fractions of 1\%, 5\%, and 10\% and 15 paired independent runs. The performance-oriented MOEA-SSL solutions achieve test MacroF1 comparable to EA-SSL in most conditions and significant gains in selected Grid Stability and Water Potability settings. More importantly, MOEA-SSL reduces median FeatureRatio relative to EA-SSL in all nine dataset--label conditions, with reductions ranging from 16.7\% to 60.0\% and statistically significant paired differences after Holm correction. A validation-based performance-retention analysis further shows that the Pareto fronts often permit additional measurement reductions with negligible changes in test MacroF1. The results support explicit Pareto optimization as a simple mechanism for exposing resource-aware SSL alternatives while preserving competitive predictive quality.

\end{abstract}

\section{Introduction}
\label{sec:introduction}

Smart cities rely on dense sensing and information infrastructures to monitor buildings, energy systems, mobility, environmental conditions, and public services. These infrastructures continuously generate heterogeneous data, but reliable labels can require manual annotation, expert inspection, intrusive measurements, or costly field procedures. Label scarcity is therefore a relevant limitation for urban machine learning systems, even when large quantities of raw observations are available~\cite{batty2012smart,mena2024occupancy,zhang2024urbanpower}.

Semi-supervised learning (SSL) addresses this setting by combining a small labeled set with a larger collection of unlabeled observations~\cite{chapelle2006ssl,vanengelen2020survey}. Lightweight approaches such as self-training and co-training suit tabular urban data because they can reuse conventional classifiers, but their performance depends on interacting decisions concerning feature views, confidence thresholds, acceptance caps, and stopping criteria~\cite{triguero2015selflabeled,arazo2020pseudo}. %These choices become especially sensitive when only 1--10\% of the training data are labeled.

Previous work introduced evolutionary formulations for adapting two-view construction and pseudo-labeling policies in tabular SSL (EA-SSL)~\cite{toutouh2026eassl}. The EA-SSL optimizes the SSL configuration and achieved competitive results against lightweight SSL baselines on 25 OpenML datasets. However, the search is scalar and returns a single configuration according to predictive fitness. In a smart-city deployment, predictive quality is only one concern: every input variable may represent a measurement, sensing modality, communication field, or preprocessing step. Maximizing classification quality and minimizing the required measurement set define a natural trade-off. Evolutionary multi-objective feature selection addresses the analogous conflict between predictive performance and subset size~\cite{jiao2024survey,xue2016survey}.

This article proposes MOEA-SSL, a multi-objective extension of EA-SSL for resource-aware smart-city classification. MOEA-SSL reformulates the evolutionary optimization layer as a bi-objective problem and applies the Non-dominated Sorting Genetic Algorithm version II (NSGA-II)~\cite{deb2002nsga2} to optimize predictive quality and feature usage separately while preserving the underlying EA-SSL representation and SSL mechanism.
The empirical study considers three classification Smart City datasets, covering smart buildings, smart energy, and smart-water monitoring. 

The main contributions of this article are: \textit{(i)} a bi-objective formulation of evolutionary SSL that separates robust predictive quality and feature usage; \textit{(ii)} MOEA-SSL, an NSGA-II extension of EA-SSL; (iii) a cross-domain study on three  Smart City classification datasets under severe label scarcity; and \textit{(iv)} a resource-oriented analysis of Pareto fronts, including performance-oriented and performance-retention operating points. 

The article is organized as follows. Section~\ref{sec:problem} formulates the multi-objective SSL optimization problem. Section~\ref{sec:related} reviews related work. Section~\ref{sec:method} presents MOEA-SSL. Section~\ref{sec:experimental} describes the experimental settings. Section~\ref{sec:analysis} analyzes the results. Finally,  Section~\ref{sec:conclusion} presents the main conclusions and future work.

\section{Multi-Objective SSL Problem Description}
\label{sec:problem}

The problem addressed in this work extends the mono-objective evolutionary SSL formulation introduced by the authors of EA-SSL~\cite{toutouh2026eassl}. That formulation jointly optimized two feature views and the pseudo-labeling policy through a scalar fitness function. Here, the same general SSL decision space is reformulated as a bi-objective problem that explicitly considers robust predictive quality and feature usage.

Let $D=D_L\cup D_U$ denote the SSL training pool, with $D_L\cap D_U=\emptyset$. The labeled subset is
$D_L=\{(\mathbf{x}_i,y_i)\}_{i=1}^{n_L}$,
and the unlabeled subset is
$D_U=\{\mathbf{x}_j\}_{j=1}^{n_U}$.
Each input vector $\mathbf{x}\in\mathbb{R}^{d}$ is described by $d$ available features, and $y\in\mathcal{Y}=\{1,\ldots,C\}$, where $C$ denotes the number of classes. The labeled fraction is
$\rho=\frac{n_L}{n_L+n_U}$.

\sloppy
The SSL procedure operates on two feature views defined by binary masks \hbox{$\mathbf{m}^{(1)},\mathbf{m}^{(2)}\in\{0,1\}^{d}$}, where $m_k^{(v)}=1$ indicates that feature $k$ is included in view $v\in\{1,2\}$. The corresponding selected-feature sets are
$S^{(v)}=\left\{k\in\{1,\ldots,d\}:m_k^{(v)}=1\right\}$,
$\qquad v\in\{1,2\}$. The masks are evolved independently and may overlap.

A candidate solution $\mathbf{z}\in\Omega$ encodes a complete SSL configuration, including the two feature masks and the pseudo-labeling parameters described in Section~\ref{sec:method}. Each candidate is evaluated over $R$ fixed labeled-data resamples. Let $F_r(\mathbf{z})$ denote validation MacroF1 for resample $r\in\{1,\ldots,R\}$. The mean $\mu_F(\mathbf{z})$ and standard deviation $\sigma_F(\mathbf{z})$ are
\begin{equation}
\begin{aligned}
\mu_F(\mathbf{z})
&=
\frac{1}{R}\sum_{r=1}^{R}F_r(\mathbf{z}),
&
\sigma_F(\mathbf{z})
&=
\sqrt{
\frac{1}{R}\sum_{r=1}^{R}
\left(F_r(\mathbf{z})-\mu_F(\mathbf{z})\right)^2
}.
\end{aligned}
\end{equation}
The robust score is defined according to $\mathrm{RobustMacroF1}(\mathbf{z})$:
\begin{equation}
\label{eq:robustf1}
\mathrm{RobustMacroF1}(\mathbf{z})=
\mu_F(\mathbf{z})-\alpha\sigma_F(\mathbf{z}),
\qquad \alpha=0.4,
\end{equation}
where $\alpha$ controls the penalty assigned to variability across low-label resamples. The value $\alpha=0.4$ follows the robustness weight used in the previous evolutionary SSL study~\cite{toutouh2026eassl}. Since both objectives are expressed as minimization criteria, the first objective is
\begin{equation}
\label{eq:f1}
f_1(\mathbf{z})=-\mathrm{RobustMacroF1}(\mathbf{z}).
\end{equation}

The second objective measures the fraction of distinct original variables required by at least one view:
\begin{equation}
\label{eq:fr}
f_2(\mathbf{z})=\mathrm{FeatureRatio}(\mathbf{z})=
\frac{|S^{(1)}\cup S^{(2)}|}{d}.
\end{equation}
The union avoids double-counting variables selected by both views. Thus, FeatureRatio measures the proportion of input information required by the SSL configuration. It is a hardware-agnostic proxy for measurement and information-acquisition complexity. % and does not assume identical monetary, sensing, communication, or energy cost for every variable.

The resulting multi-objective optimization problem is
\begin{equation}
\label{eq:biobjective}
\min_{\mathbf{z}\in\Omega}
\left(f_1(\mathbf{z}),f_2(\mathbf{z})\right).
\end{equation}
Given two feasible configurations $\mathbf{z}_a,\mathbf{z}_b\in\Omega$, $\mathbf{z}_a$ dominates $\mathbf{z}_b$ if $f_i(\mathbf{z}_a)\leq f_i(\mathbf{z}_b)$ for $i\in{1,2}$ and the inequality is strict for at least one objective. The aim is to obtain a set of Pareto-non-dominated SSL configurations representing trade-offs between robust predictive performance and feature usage, rather than a single weighted compromise.

\section{Related Work}
\label{sec:related}

SSL combines labeled and unlabeled data to improve learning when annotation is limited~\cite{chapelle2006ssl,vanengelen2020survey}. Self-training iteratively augments the labeled set with confident predictions, while co-training exchanges pseudo-labels between learners trained on different views~\cite{blum1998cotraining,triguero2015selflabeled}. Their simplicity is attractive for tabular data, although performance can deteriorate when confidence estimates are unreliable or erroneous pseudo-labels are reinforced~\cite{arazo2020pseudo}. Neural tabular methods such as VIME extend self-supervised and semi-supervised representation learning to this domain~\cite{yoon2020vime}, but require a more complex training pipeline.

Evolutionary and coevolutionary learning has also been applied directly to SSL. Building on spatial evolutionary GAN training~\cite{toutouh2019mustangs}, Toutouh et al.~\cite{toutouh2023lipissl,toutouh2023coessl} combined semi-supervised GANs with spatial coevolution, maintaining populations of generators and discriminators in a distributed grid and exchanging components according to performance. These studies focus on image-based adversarial deep learning and use coevolution as the training mechanism. In contrast, the present work addresses lightweight tabular SSL and uses evolutionary search to configure a pseudo-labeling process rather than to train the predictive model parameters directly.

Label scarcity is also documented in smart-city sensing. Mena-Martinez et al.~\cite{mena2024occupancy} evaluated supervised and semi-supervised methods for occupancy estimation from environmental IoT data. Ghourchian et al.~\cite{ghourchian2017smart} applied SSL to WiFi-based indoor localization with limited user annotation, and Zhang et al.~\cite{zhang2024urbanpower} proposed adaptive semi-supervised fault classification for urban power grids. These works demonstrate the relevance of SSL in buildings and energy systems, but they primarily optimize predictive performance rather than an explicit performance--measurement trade-off.

Feature selection is a combinatorial problem for which evolutionary computation has been extensively studied~\cite{song2024efs,toutouh2018bipred,xue2016survey}. Multi-objective formulations commonly maximize predictive performance while minimizing subset size, returning alternative trade-offs for different decision requirements~\cite{jiao2024survey}.

MOEAs are widely used in smart-city decision support because urban problems naturally combine conflicting service, economic, environmental, and infrastructure criteria~\cite{chen2025moo}. Previous applications include the optimization of competing communication criteria in vehicular networks~\cite{toutouh2017vehicular}, multi-objective location of garbage accumulation points~\cite{toutouh2020garbage}, household energy planning balancing energy use and user preferences~\cite{colacurcio2019energy}, electric-vehicle charging-station location trading quality of service against deployment cost~\cite{cintrano2022evcs}, and e-scooter infrastructure redesign considering mobility quality and installation cost~\cite{pedroza2024escooter}. Related transportation studies have also applied Pareto-based evolutionary search to multimodal route finding~\cite{faroqi2024route} and transit-network design~\cite{chai2020transit}. These works illustrate the value of Pareto fronts for exposing alternative urban trade-offs instead of collapsing conflicting criteria into a single weighted objective. The present study transfers this decision-support perspective to SSL configuration, where predictive robustness conflicts with the amount of measurement information required by the classifier.

The closest antecedent is EA-SSL~\cite{toutouh2026eassl}, which jointly evolves two feature-subset views and a pseudo-labeling policy. Its scalar formulation, however, does not explicitly distinguish predictive robustness from feature usage. MOEA-SSL addresses this gap by preserving the EA-SSL search space and SSL mechanism while optimizing robust MacroF1 and FeatureRatio as separate objectives.

\section{MOEA-SSL Method}
\label{sec:method}

MOEA-SSL extends the monolithic EA-SSL method~\cite{toutouh2026eassl}
without changing its SSL representation or pseudo-labeling mechanism.
%Specifically, it retains the EA-SSL genotype, two-view construction, regularized linear base learner, pseudo-labeling procedure, variation operators, and feasibility repairs.
The methodological change concerns the evolutionary formulation: the original scalar fitness is replaced by the separate predictive-performance and feature-usage objectives defined in Section~\ref{sec:problem}, which are optimized through NSGA-II~\cite{deb2002nsga2}.

\subsection{Representation and SSL evaluation}
\label{subsec:representation}

A candidate is
represented as $\mathbf{z}=(\mathbf{a},\mathbf{b})$,
where $\mathbf{a}$ is the view-builder component and $\mathbf{b}$ is the
pseudo-labeling-policy component. The view builder is
\begin{equation}
\mathbf{a}=
\left(
\mathbf{m}^{(1)},\mathbf{m}^{(2)},
p^{(1)},p^{(2)},
k^{(1)},k^{(2)},
B^{(1)},B^{(2)}
\right),
\label{eq:viewgenotype}
\end{equation}
where $\mathbf{m}^{(v)}\in\{0,1\}^{d}$ selects the features of view $v$,
$p^{(v)}\in\{0,1\}$ activates the optional projection,
$k^{(v)}\in\{2,\ldots,k_{\max}\}$ specifies its dimension, and
$B^{(v)}\in\{0,\ldots,B_{\max}\}$ controls discretization. The two masks are
evolved independently and may overlap.

The policy component is
\begin{equation}
\mathbf{b}=
\left(
\boldsymbol{\theta}_{\mathrm{clf}},
\tau_0,\Delta\tau,\tau_{\min},
q,\gamma,\nu,T
\right),
\label{eq:policygenotype}
\end{equation}
where $\boldsymbol{\theta}_{\mathrm{clf}}$ contains the hyperparameters of
the regularized linear classifier used by EA-SSL (logistic regression),
$\tau_0\in(0,1)$ is the initial confidence threshold,
$\Delta\tau\geq0$ its per-iteration decay,
$\tau_{\min}\in(0,1)$ its lower bound, $q\in\mathbb{N}$ the per-class
pseudo-label acceptance cap, $\gamma\in[0,1]$ the minimum posterior margin,
$\nu\in\{0,1\}$ the disagreement-veto flag, and $T\in\mathbb{N}$ the maximum
number of pseudo-labeling iterations.

Candidate evaluation uses the inherited two-view SSL procedure. Starting
from $D_L$ and $D_U$, one classifier is trained on each evolved view at every
iteration. The confidence threshold follows
\begin{equation}
\tau_t=\max\{\tau_{\min},\tau_0-t\Delta\tau\}.
\label{eq:tauschedule}
\end{equation}
Predictions on the unlabeled pool are filtered by the confidence threshold,
the optional posterior-margin requirement, the optional cross-view
disagreement veto, and the per-class acceptance cap. Accepted observations
are transferred from the unlabeled to the labeled pool. Iteration terminates
when no further observations are accepted or after $T$ iterations, and the
resulting SSL configuration is scored on the validation partition.

The SSL evaluation is repeated over the $R$ fixed low-label resamples.
Their validation MacroF1 values define $f_1(\mathbf{z})$ through
Eqs.~\eqref{eq:robustf1}--\eqref{eq:f1}. The coefficient $\alpha=0.4$
inherits the standard-deviation robustness weight used in the previous
evolutionary SSL study~\cite{toutouh2026eassl}. This objective corresponds
to the mean--variability performance component of the previous scalar
fitness; the original formulation also contained auxiliary terms and is
therefore not identical to the MOEA-SSL predictive objective.

The second objective is obtained directly from the evolved masks:
\begin{equation}
f_2(\mathbf{z})=\mathrm{FeatureRatio}(\mathbf{z})
=\frac{|S^{(1)}\cup S^{(2)}|}{d},
\end{equation}
as defined in Eq.~\eqref{eq:fr}. The union prevents double-counting a
feature selected by both views. Hence, FeatureRatio is available directly
from the genotype and requires no additional classifier-training evaluation.

\subsection{Multi-objective evolutionary search}
\label{subsec:nsgasearch}

MOEA-SSL evolves a population of complete monolithic EA-SSL configurations
using NSGA-II. A feasible population is randomly initialized and evaluated
on $(f_1,f_2)$. Parent selection uses binary tournament based on
non-dominated rank and crowding distance. Offspring are generated with the
inherited EA-SSL variation operators, repaired, and evaluated on the same
fixed resamples. Parent and offspring populations are then merged, and
non-dominated sorting and crowding distance determine environmental
survival. Algorithm~\ref{alg:moeassl} summarizes the procedure.

\begin{algorithm}[!h]
\caption{MOEA-SSL}
\label{alg:moeassl}
\begin{algorithmic}[1]
\Require $D_L,D_U$, validation set $V$, population size $N$,
generations $G$, fixed resamples $R$
\Ensure Non-dominated set $P^{\star}$

\State Initialize and repair population $P_0$ of $N$ candidates
\State Evaluate $(f_1,f_2)$ for every $\mathbf{z}\in P_0$

\For{$g=0,\ldots,G-1$}
    \State Rank $P_g$ by non-dominated sorting and crowding distance
    \State $O_g\gets\emptyset$

    \While{$|O_g|<N$}
        \State Select parents by rank--crowding binary tournament
        \State Apply inherited EA-SSL crossover and mutation
        \State Repair and evaluate offspring
        \State Add offspring to $O_g$
    \EndWhile

    \State $M_g\gets P_g\cup O_g$
    \State Compute Pareto fronts of $M_g$
    \State Fill $P_{g+1}$ front by front
    \State Truncate the last front by crowding distance if necessary
\EndFor

\State $P^{\star}\gets\mathcal{F}_1(P_G)$
\State \Return $P^{\star}$
\end{algorithmic}
\end{algorithm}

The returned $P^{\star}$ contains alternative validation-non-dominated
trade-offs between RobustMacroF1 and FeatureRatio. Test data are not used
during evolution or validation-based operating-point selection. Any selected
configuration, including the performance-oriented solution defined in
Section~\ref{sec:experimental}, is evaluated on the untouched test partition
only after validation-based selection.

\subsection{Evolutionary operators}
\label{subsec:operators}

MOEA-SSL retains the EA-SSL variation operators unchanged and applies them
according to gene type. Table~\ref{tab:operators} summarizes the operators;
variation is applied independently to the view-builder and policy blocks
using the probabilities reported in Section~\ref{sec:experimental}.

\begin{table}[!h]
\caption{Variation operators inherited from EA-SSL.}
\label{tab:operators}
\centering
\small
\begin{tabular}{p{3.0cm}p{3.6cm}p{3.7cm}}
\toprule
Component & Crossover & Mutation \\
\midrule
Feature masks &
Uniform exchange &
Bit flip \\

Boolean genes &
Uniform exchange &
Boolean flip \\

Integer genes &
Uniform exchange &
Bounded $\pm1$ step \\

Continuous genes &
Arithmetic crossover &
Bounded perturbation \\
\bottomrule
\end{tabular}
\end{table}

Repair is performed before evaluation. For each view, the inherited
minimum-feature constraint $\|\mathbf{m}^{(v)}\|_0 \geq k_{\min},
\qquad v\in\{1,2\}$,
is enforced, and all genes are kept within their admissible domains. Policy
repair additionally ensures
$\tau_{\min}\leq\tau_0$ and $\Delta\tau\geq0$.
These repairs use no predictive information.

\section{Experimental Settings}
\label{sec:experimental}

The experimental study considers three binary Smart City classification datasets covering building operation, smart energy, and water-quality monitoring: Building Occupancy~\cite{uciOccupancy}, Grid Stability~\cite{uciGrid}, and Water Potability~\cite{kadiwalWater}. Table~\ref{tab:datasets} summarizes their main characteristics.

\begin{table}[!h]
\setlength{\tabcolsep}{0.1cm}
\caption{Datasets used in the experimental evaluation.}
\label{tab:datasets}
\centering
\small
\begin{tabular}{l l r r }
\toprule
Dataset & Domain & Samples & Number of features, $d$ \\
\midrule
Building Occupancy & Smart buildings & 10,808 & 7  \\
Grid Stability & Smart energy & 10,000 & 12 \\
Water Potability & Smart water & 3,276 & 9 \\
\bottomrule
\end{tabular}
\end{table}

Building Occupancy uses the repository training partition and first held-out
test partition, yielding 10,808 observations and seven processed predictors.
Grid Stability uses 12 predictors for the categorical stable/unstable target;
the continuous target \texttt{stab} is excluded. The redundant input $p_1$
is retained because feature redundancy is part of the optimization problem.
Water Potability contains nine physicochemical predictors; missing values are
median-imputed from training data.

For Building Occupancy, the existing held-out test partition contains 2,665 samples and 20\% of the repository training partition is used for validation (1,629 samples). For Grid Stability and Water Potability, 30\% of the data are held out for testing and 20\% of the remaining development partition is used for validation. The remaining development data form $D_L\cup D_U$. For each random seed, $D_L$ is obtained using $\rho\in\{0.01,0.05,0.10\}$ and the remainder constitutes $D_U$. The original Building Occupancy train/test organization is preserved, whereas Grid Stability and Water Potability use stratified random partitions. Scaling and imputation parameters are fitted without access to validation or test labels.

\subsection{Baselines and parameter settings}

The analysis uses two lightweight SSL baselines from the previous
study~\cite{toutouh2026eassl}: Self-Training (ST) and Heuristic Co-Training
(HCo). Label Spreading is excluded because its graph construction exceeded
the adopted size limit for Building Occupancy and Grid Stability. Four
classifiers trained only on $D_L$ are reported as descriptive supervised
references: Logistic Regression (LR), linear SVM, Random Forest (RF), and
Histogram-based Gradient Boosting (HGB). The primary evolutionary comparator
is EA-SSL~\cite{toutouh2026eassl}.

EA-SSL uses the same representation, SSL procedure, variation operators, and
evolutionary budget as MOEA-SSL, while retaining its original scalar fitness.
Both methods use a population of 36, 50 generations, and $R$=$10$ fixed
low-label resamples for candidate evaluation. Crossover and mutation
probabilities are 0.85 and 0.35 for both the view and policy components,
following~\cite{toutouh2026eassl}. Each method is independently executed
with 30 shared random seeds per dataset and labeled fraction, enabling paired
statistical comparisons.

\subsection{Evaluation methodology}

MacroF1 is the primary predictive metric because it assigns equal weight to every class. Classification accuracy is retained as a secondary diagnostic. For scalar comparison against EA-SSL and the baselines, one MOEA-SSL configuration is selected from each validation front without using test labels:
\begin{equation}
\label{eq:perfsolution}
\mathbf{z}_{\mathrm{perf}}=
\arg\max_{\mathbf{z}\in P_V}
\mathrm{RobustMacroF1}_V(\mathbf{z}),
\end{equation}
where ties are broken by lower FeatureRatio. The selected configuration is then evaluated on the untouched test set. This performance-oriented point provides a conservative comparison with the accuracy-oriented EA-SSL solution while the complete front is retained for resource-trade-off analysis.

To quantify whether the front offers useful alternatives beyond this extreme, a second validation-only operating point is defined. Let $F_V^{\max}=\mathrm{RobustMacroF1}_V(\mathbf{z}_{\mathrm{perf}})$. The one-point performance-retention solution is
\begin{equation}
\label{eq:retention}
\mathbf{z}_{1\mathrm{pp}}=
\arg\min_{\mathbf{z}\in P_V}\mathrm{FeatureRatio}(\mathbf{z})
\quad\text{s.t.}\quad
\mathrm{RobustMacroF1}_V(\mathbf{z})\geq F_V^{\max}-0.01.
\end{equation}
Thus, $\mathbf{z}_{1\mathrm{pp}}$ accepts at most one absolute MacroF1 point of validation degradation in exchange for a smaller measurement set. Test values are again inspected only after selection. This analysis is descriptive and does not add further hypothesis tests.

Multi-objective quality is characterized using FeatureRatio, the number of distinct selected variables, representative Pareto sets, and hypervolume (HV)~\cite{guerreiro2021hypervolume}. HV is computed on the validation objectives
\[
(1-\mathrm{RobustMacroF1},\,\mathrm{FeatureRatio}),
\]
using the fixed reference point $(1,1)$. Representative fronts are chosen as the run whose HV is closest to the median HV for the corresponding dataset and labeled fraction. Their non-dominated validation solutions are subsequently projected onto test MacroF1 for visualization. Test values are never used to construct or choose the validation front.

Descriptive results are reported as median and interquartile range (IQR) over the 30 shared seeds. Paired comparisons use the Wilcoxon signed-rank test with significance level $\alpha_s=0.05$~\cite{demsar2006statistical}. For predictive performance, Holm correction is applied to the three comparisons of MOEA-SSL against EA-SSL, ST, and HCo within each dataset--fraction condition. FeatureRatio constitutes a pre-specified direct MOEA-SSL versus EA-SSL comparison. The nine corresponding tests are Holm-corrected jointly. Supervised methods are used as descriptive references and are not included in the multiple-comparison tests.

\subsection{Execution platform and implementation}
\label{subsec:platform}
The methods were implemented in Python 3 mainly using NumPy, pandas, and scikit-learn   . Experiments were executed on a high-performance cluster with 126 SD530 nodes, each equipped with 52 Intel Xeon Gold 6230R cores at 2.10\,GHz and 192\,GB of RAM, interconnected through InfiniBand HDR100. The implementation and experimental scripts are available at the project repository.\footnote{\url{https://github.com/jamaltoutouh/moea-ssl-smat-cities}}

\section{Experimental Analysis}
\label{sec:analysis}

This section evaluates MOEA-SSL from complementary perspectives. First, the
performance-oriented validation solution $\mathbf{z}_{\mathrm{perf}}$ is
compared with EA-SSL and the lightweight baselines on the held-out test set.
Second, the non-dominated sets are analyzed in terms of feature efficiency
and resource--performance trade-offs. Third, $\mathbf{z}_{1\mathrm{pp}}$
examines the additional compression obtained when a small validation-performance
concession is allowed. Finally, the influence of labeled-data availability,
computational cost, and the main limitations are discussed.

\subsection{Predictive performance}
\label{subsec:predictive}

Table~\ref{tab:macro} reports test MacroF1. Building Occupancy presents a
clear ceiling effect: EA-SSL and MOEA-SSL remain close to 0.977 at every
labeled fraction, with no significant paired differences. MOEA-SSL
nevertheless significantly improves over HCo in all three conditions and
over ST at 1\% after within-condition Holm correction.

\begin{table}[!h]
\caption{Test MacroF1 across 30 runs. Values are median[inter-quartile range]. Bold indicates the highest median among the SSL methods. Best Sup. is the highest-median supervised approach and is descriptive only. $\dagger$ denotes a significant MOEA-SSL vs. EA-SSL difference (Holm correction).}
\label{tab:macro}
\centering
%\scriptsize
\renewcommand{\arraystretch}{0.9}
\small
\setlength{\tabcolsep}{2.2pt}
\begin{tabular}{l rrrrrr}
\toprule
Dataset & $\rho$ & ST & HCo & EA-SSL & MOEA-SSL & Best Sup. \\
\midrule
Building & 1\% & 0.950[0.055] & 0.881[0.040] & 0.977[0.000] & \textbf{0.977[0.000]} & 0.977[0.005] (SVM) \\
 & 5\% & 0.977[0.000] & 0.947[0.014] & \textbf{0.977[0.000]} & 0.977[0.000] & 0.978[0.000] (SVM) \\
 & 10\% & 0.977[0.000] & 0.948[0.014] & 0.977[0.000] & \textbf{0.977[0.001]} & 0.978[0.001] (SVM) \\
\midrule
Grid & 1\% & 0.683[0.076] & 0.660[0.166] & 0.746[0.032] & \textbf{0.755[0.021]} & 0.737[0.040] (SVM) \\
 & 5\% & 0.757[0.020] & 0.533[0.174] & 0.764[0.017] & \textbf{0.770[0.010]} & 0.829[0.016] (HGB) \\
 & 10\% & \textbf{0.782[0.007]} & 0.728[0.026] & 0.761[0.010] & 0.777[0.009]$^{\dagger}$ & 0.862[0.013] (HGB) \\
\midrule
Water & 1\% & 0.424[0.066] & 0.379[0.011] & 0.491[0.031] & \textbf{0.500[0.011]}$^{\dagger}$ & 0.495[0.028] (SVM) \\
 & 5\% & 0.383[0.038] & 0.379[0.010] & 0.471[0.027] & \textbf{0.506[0.022]}$^{\dagger}$ & 0.517[0.029] (HGB) \\
 & 10\% & 0.385[0.011] & 0.379[0.003] & 0.439[0.039] & \textbf{0.471[0.071]} & 0.536[0.016] (HGB) \\
\bottomrule
\end{tabular}
\end{table}

Grid Stability provides a more discriminative comparison. MOEA-SSL obtains
median MacroF1 values of 0.755, 0.770, and 0.777 for 1\%, 5\%, and 10\%
labeled data, compared with 0.746, 0.764, and 0.761 for EA-SSL. The
difference is significant at 10\% ($p_{\mathrm{Holm}}=0.0057$). The
supervised references also become increasingly competitive as more labels
are available: HGB reaches 0.829 and 0.862 at 5\% and 10\%, respectively.

Water Potability shows the clearest predictive advantage for MOEA-SSL. The
EA-SSL median increases from 0.491 to 0.500 at 1\% labels and from 0.471 to
0.506 at 5\%, with significant differences after correction
($p_{\mathrm{Holm}}=0.035$ and $0.015$, respectively). At 10\%, MOEA-SSL
retains a higher median (0.471 vs.\ 0.439), although the difference is not
significant. MOEA-SSL also significantly outperforms ST and HCo at all three
labeled fractions, while HGB becomes the strongest descriptive supervised
reference at the larger fractions.

Overall, MOEA-SSL is competitive rather than uniformly superior in predictive
performance. It remains close to EA-SSL on Building and most Grid/Water
conditions, while providing significant gains in selected Grid and Water
settings. More importantly, the proposed bi-objective formulation preserves
competitive predictive quality while enabling substantially smaller feature
sets.

\subsection{Feature efficiency and Pareto trade-offs}
\label{subsec:tradeoff}

Table~\ref{tab:feature} reports FeatureRatio for the same performance-oriented
MOEA-SSL solutions used in Table~\ref{tab:macro}. Despite selecting the
validation-performance extreme of each Pareto set, MOEA-SSL requires fewer
distinct variables than EA-SSL in every dataset--fraction condition. The
reduction ranges from 16.7\% to 60.0\%, and all nine paired differences remain
significant after Holm correction across the nine pre-specified FeatureRatio
tests.

\begin{table}[!h]
\caption{Feature usage of the performance-oriented MOEA-SSL solution and EA-SSL. Values in parentheses are median selected features over $d$. Reduction is computed from the median FeatureRatio values. All paired FeatureRatio differences remain significant after Holm correction.}
\label{tab:feature}
\centering
%\scriptsize
\setlength{\tabcolsep}{4.0pt}
\renewcommand{\arraystretch}{0.9}
\small
\begin{tabular}{l rrrrr}
\toprule
Dataset & $\rho$ & MOEA-SSL & EA-SSL & Reduction & $p_{\mathrm{Holm}}$ \\
\midrule
Building & 1\% & 0.429 (3/7) & 0.857 (6/7) & 50.0\% & 0.005 \\
 & 5\% & 0.429 (3/7) & 0.714 (5/7) & 40.0\% & 0.010 \\
 & 10\% & 0.286 (2/7) & 0.714 (5/7) & 60.0\% & 0.007 \\
\midrule
Grid & 1\% & 0.667 (8/12) & 0.833 (10/12) & 20.0\% & 0.011 \\
 & 5\% & 0.667 (8/12) & 0.833 (10/12) & 20.0\% & 0.010 \\
 & 10\% & 0.750 (9/12) & 0.917 (11/12) & 18.2\% & 0.014 \\
\midrule
Water & 1\% & 0.444 (4/9) & 0.667 (6/9) & 33.3\% & 0.045 \\
 & 5\% & 0.556 (5/9) & 0.667 (6/9) & 16.7\% & 0.045 \\
 & 10\% & 0.556 (5/9) & 0.778 (7/9) & 28.6\% & 0.013 \\
\bottomrule
\end{tabular}
\end{table}

The reduction varies across problems. In Building Occupancy, the median
performance-oriented MOEA solution uses two or three of seven variables,
compared with five or six for EA-SSL. Grid Stability shows a smaller but
consistent reduction from approximately 10--11 of 12 variables to 8--9.
Water Potability reduces the median measurement set from 6--7 of 9 variables
to 4--5. Thus, the most consistent effect of the second objective is a shift
toward substantially more compact input representations.

\begin{figure}[!h]
\centering
\includegraphics[width=\textwidth]{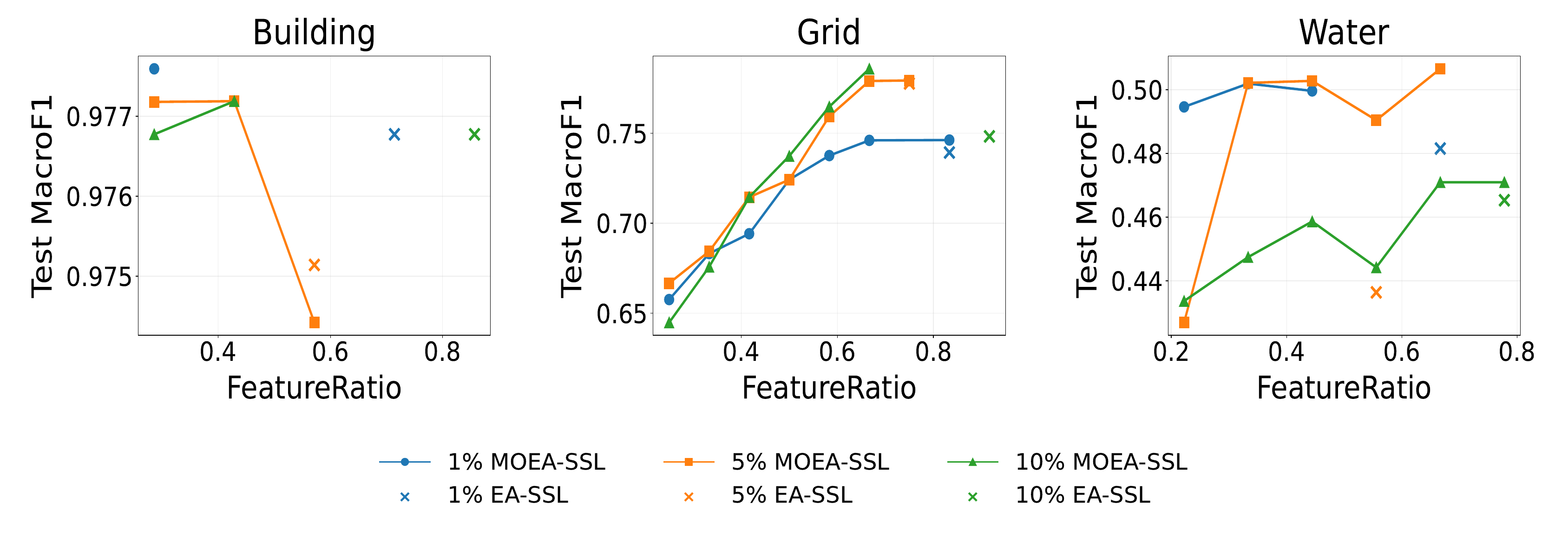}
\caption{Test-space projection of validation-non-dominated solutions for representative median-HV runs. MOEA-SSL sets are selected using validation objectives only (vs. same sed EA-SSL).}
\label{fig:tradeoff}
\end{figure}

Fig.~\ref{fig:tradeoff} illustrates the
test-space projection of representative validation fronts. For each dataset
and labeled fraction, the plotted MOEA-SSL set corresponds to the run whose
HV is closest to the median. Building Occupancy shows that sparse configurations can preserve
near-ceiling performance. Grid Stability presents the clearest smooth
quality--feature trade-off, with intermediate solutions unavailable from the
scalar EA-SSL output. Water Potability shows a less regular test projection,
particularly at 10\%, illustrating that validation non-dominance does not
guarantee test non-dominance. The median number of distinct non-dominated
objective vectors is two for Building, six to seven for Grid, and four for
Water, showing that the richness of the trade-off is problem dependent.

These results show that optimizing FeatureRatio already produces substantially
smaller measurement sets at the performance-oriented extreme. The following
analysis examines whether further compression is available elsewhere on the
front.

\subsection{Performance-retention operating points}
\label{subsec:retention}

Table~\ref{tab:retention} evaluates the validation-only rule in
Eq.~\eqref{eq:retention}. The median feature count decreases beyond
$\mathbf{z}_{\mathrm{perf}}$ in six of the nine conditions. Building reduces
from three to two variables at 1\% and 5\% labels while preserving a median
test MacroF1 of approximately 0.977. Grid removes one additional variable at
1\% and 10\%; the largest median test decrease is only 0.008 at Grid 1\%.
Water provides the strongest additional compression at 10\%, where the median
measurement set falls from five to three of nine variables while test MacroF1
changes from 0.471 to 0.469.

\begin{table}[!h]
\caption{Performance-retention analysis. $\mathbf{z}_{1\mathrm{pp}}$ minimizes FeatureRatio within 0.01 validation RobustMacroF1 of $\mathbf{z}_{\mathrm{perf}}$. Feature counts and test MacroF1 are medians over 30 runs; ``Reduced'' counts runs in which $\mathbf{z}_{1\mathrm{pp}}$ uses fewer features.}
\label{tab:retention}
\centering
\small
\setlength{\tabcolsep}{3.1pt}
\begin{tabular}{l c c c c c c}
\toprule
Dataset & $\rho$ & $z_{\mathrm{perf}}$ feat. & $z_{1\mathrm{pp}}$ feat. & Reduced & $F1_{\mathrm{perf}}$ & $F1_{1\mathrm{pp}}$ \\
\midrule
Building & 1\% & 3/7 & 2/7 & 16/30 & 0.9772 & 0.9772 \\
 & 5\% & 3/7 & 2/7 & 16/30 & 0.9768 & 0.9772 \\
 & 10\% & 2/7 & 2/7 & 18/30 & 0.9772 & 0.9772 \\
\midrule
Grid & 1\% & 8/12 & 7/12 & 22/30 & 0.7554 & 0.7479 \\
 & 5\% & 8/12 & 8/12 & 12/30 & 0.7701 & 0.7660 \\
 & 10\% & 9/12 & 8/12 & 20/30 & 0.7770 & 0.7788 \\
\midrule
Water & 1\% & 4/9 & 4/9 & 16/30 & 0.4996 & 0.5000 \\
 & 5\% & 5/9 & 4/9 & 20/30 & 0.5058 & 0.5028 \\
 & 10\% & 5/9 & 3/9 & 20/30 & 0.4709 & 0.4687 \\
\bottomrule
\end{tabular}
\end{table}

Even when the median feature count is unchanged, the retention rule finds
smaller solutions in multiple paired runs. Test changes remain small because
selection is constrained on validation rather than optimized on test data.
Hence, the benefit of the multi-objective formulation extends beyond making
the performance extreme more compact than EA-SSL: the Pareto front also
provides deployment alternatives that exchange a bounded validation-performance
concession for further reductions in measurement requirements.

\subsection{Influence of labeled-data availability}
\label{subsec:labels}

Increasing the labeled fraction does not produce a universal monotonic effect
on MOEA-SSL. Grid Stability improves progressively as $\rho$ increases,
whereas Building Occupancy remains close to its performance ceiling.
Water Potability reaches its highest median test MacroF1 at 5\% labeled data
and decreases at 10\%. Feature usage is also problem dependent: Building
becomes more compact at 10\%, Grid uses slightly more variables, and Water
increases from four to five median variables after 1\%.

Water Potability also shows a marked adaptation in pseudo-labeling behavior.
The median number of accepted pseudo-labels decreases from 852 at 1\% labeled
data to 466 at 5\% and only one at 10\%. Thus, as more labeled information
becomes available, evolution can favor configurations that largely suppress
pseudo-labeling rather than forcing additional unlabeled observations into
training. Performance in this regime should therefore be interpreted as the
result of jointly optimizing the feature views, classifier configuration, and
SSL policy, rather than being attributed solely to pseudo-label augmentation.

\subsection{Multi-objective quality and computational cost}
\label{subsec:qualitycost}

Table~\ref{tab:hvruntime} reports the validation hypervolume and wall-clock
runtime of MOEA-SSL and EA-SSL. Hypervolume is relatively stable across
labeled fractions for Building Occupancy and Grid Stability. Building ranges
from 0.701 to 0.706, while Grid remains between 0.565 and 0.573. Water
Potability exhibits a different pattern, with median HV decreasing from
0.456 at 1\% labeled data to 0.407 at 10\%. These values characterize the
quality and coverage of the obtained validation Pareto sets and should not
be interpreted as direct measures of classification-task difficulty.

\begin{table}[!h]
\caption{Validation hypervolume and computational cost. Values are
median[IQR] over the independent runs. Runtime is reported in seconds and
Ratio denotes the median MOEA-SSL runtime divided by the median EA-SSL
runtime.}
\label{tab:hvruntime}
\centering
\small
\setlength{\tabcolsep}{3.2pt}
\begin{tabular}{l c c r r c}
\toprule
Dataset & $\rho$ & HV & MOEA-SSL & EA-SSL & Ratio \\
\midrule
Building & 1\%  & 0.701[0.100] & 2487[1300] & 1618[237] & 1.54 \\
         & 5\%  & 0.705[0.138] & 2847[1783] & 1492[331] & 1.91 \\
         & 10\% & 0.706[0.139] & 3329[1950] & 1610[302] & 2.07 \\
\midrule
Grid     & 1\%  & 0.568[0.071] & 1177[344]  & 1183[283] & 1.00 \\
         & 5\%  & 0.573[0.040] & 2434[686]  & 1392[269] & 1.75 \\
         & 10\% & 0.565[0.054] & 1969[1512] & 1427[250] & 1.38 \\
\midrule
Water    & 1\%  & 0.456[0.043] & 1166[635]  & 608[141]  & 1.92 \\
         & 5\%  & 0.423[0.077] & 540[377]   & 292[74]   & 1.85 \\
         & 10\% & 0.407[0.074] & 351[219]   & 283[41]   & 1.24 \\
\bottomrule
\end{tabular}
\end{table}

Although both evolutionary methods use matched population and generation
budgets, MOEA-SSL requires more wall-clock time in most conditions. The
median runtime ratio ranges from approximately parity for Grid at 1\% to
2.07 for Building at 10\%. The overhead is therefore problem dependent
rather than constant. Since the cost of the inner SSL evaluation also depends
on the evolved pseudo-labeling behavior, these differences should be
interpreted as end-to-end computational cost rather than as the isolated
cost of NSGA-II. In return, MOEA-SSL produces a set of alternative
performance--feature trade-offs instead of the single configuration returned
by EA-SSL.

\subsection{Generalization and limitations}
\label{subsec:limitations}

Validation--test behavior is broadly consistent with the intended selection
protocol. For Building Occupancy, the median difference between validation
RobustMacroF1 and test MacroF1 of $\mathbf{z}_{\mathrm{perf}}$ remains below
0.01. The corresponding gap is approximately 0.03 for Grid Stability and
between 0.04 and 0.07 for Water Potability. These values indicate that the
validation-selected performance-oriented solutions transfer consistently
enough to support comparison on the held-out test partitions.

Three limitations are relevant. First, FeatureRatio assigns the same unit
cost to every variable and therefore approximates, rather than directly
measures, sensor, communication, storage, or energy cost. Second, the study
uses standard NSGA-II without strategic view partitioning, adaptive variation,
or problem-specific sensor costs, leaving substantial algorithmic extensions
unexplored. Third, the evaluation considers three binary, relatively
low-dimensional Smart City datasets. However, MOEA-SSL does not optimize
feature subsets in isolation: each candidate jointly defines two feature views
together with projection, discretization, classifier, and pseudo-labeling
policy decisions, yielding a richer search space than conventional feature
selection. The inherited EA-SSL framework previously reported competitive performance
across 25 OpenML datasets, including higher-dimensional problems~\cite{toutouh2026eassl};
nevertheless, validation of MOEA-SSL on higher-dimensional, multiclass, and
additional urban problems is required before drawing stronger general conclusions.

\section{Conclusions and Future Work}
\label{sec:conclusion}

This article presented MOEA-SSL, a multi-objective reformulation of EA-SSL for resource-aware semi-supervised classification in Smart City applications. The proposed method retains the same two-view pseudo-labeling representation and evolutionary operators, while replacing the original scalar fitness with two explicitly separated criteria---robust validation MacroF1 and FeatureRatio---optimized through NSGA-II.

Experiments on three Smart City datasets and three low-label regimes show that the principal benefit of the multi-objective formulation is not uniform predictive improvement, but the ability to preserve competitive performance while substantially reducing the number of required variables. The performance-oriented MOEA-SSL solution uses a significantly smaller FeatureRatio than EA-SSL in all nine dataset--fraction conditions after Holm correction, with median reductions between 16.7\% and 60.0\%. Predictive behavior remains problem dependent: MOEA-SSL is essentially tied with EA-SSL on Building Occupancy, significantly improves over EA-SSL for Grid Stability at 10\% labels and Water Potability at 1\% and 5\%, and remains competitive in the remaining conditions. Moreover, the one-point performance-retention rule further reduces the median feature count in six of nine conditions; across the reported medians, the corresponding absolute test-MacroF1 change remains below 0.008.

These results support multi-objective evolutionary search as a useful decision-support mechanism when Smart City classifiers must balance predictive quality with measurement complexity. Rather than committing to a single scalar optimum, the resulting Pareto front exposes alternative operating points with different information requirements; the performance-retention analysis demonstrates that these alternatives can remain useful even after the performance-oriented extreme has been selected. FeatureRatio should nevertheless be interpreted as a generic, hardware-agnostic proxy for input-information requirements rather than as a direct deployment-cost model.

Future work will extend the present baseline formulation in four directions. First, FeatureRatio will be replaced or complemented with explicit sensor, acquisition, communication, latency, and energy costs when such metadata are available. Second, cooperative coevolution and strategic view construction will be investigated to determine whether decomposition improves the search of larger feature spaces. Third, uncertainty-aware and adaptive multi-objective mechanisms will be studied to improve robustness under very small labeled pools and distribution shift. Finally, the evaluation will be expanded to additional cities and datasets, with particular attention to multiclass Smart City classification and broader stochastic validation.

\begin{credits}
\subsubsection{Acknowledgments}
This work has been partially funded by the AIM-Zero (PID2024-158752OB-I00) and by AGENT-SAM research projects (PID2025-174277OA-I00).
\end{credits}

% TODO before submission: add final acknowledgments and competing-interest statement.

\bibliographystyle{splncs04}
\bibliography{references}

\end{document}